\documentclass[a4paper,conference]{IEEEtran}

\usepackage{amsmath}
\usepackage{mathtools}
\usepackage{graphicx}
\usepackage{algorithm2e}
\usepackage{svg}

\DeclarePairedDelimiter\floor{\lfloor}{\rfloor}
\usepackage{diagbox}
\usepackage{todonotes}

\usepackage{array,booktabs}
\usepackage{fancyhdr}
\newcolumntype{P}[1]{>{\centering\arraybackslash}p{#1}}

\ifCLASSOPTIONcompsoc
  \usepackage[caption=false,font=normalsize,labelfont=sf,textfont=sf]{subfig}
\else
  \usepackage[caption=false,font=footnotesize]{subfig}
\fi

\usepackage{url}

\begin{document}
%
\title{Fast matrix multiplication for binary and ternary CNNs on ARM CPU}
\lhead{Accepted to ICPR 2022}



%
\author{\IEEEauthorblockN{Anton Trusov\IEEEauthorrefmark{1}\IEEEauthorrefmark{2}\IEEEauthorrefmark{3},
Elena Limonova\IEEEauthorrefmark{2}\IEEEauthorrefmark{3},
Dmitry Nikolaev\IEEEauthorrefmark{3}\IEEEauthorrefmark{4} and
Vladimir V. Arlazarov\IEEEauthorrefmark{2}\IEEEauthorrefmark{3}}
\IEEEauthorblockA{\IEEEauthorrefmark{1} Moscow Institute of Physics and Technology, Dolgoprudny, Russia}
\IEEEauthorblockA{\IEEEauthorrefmark{2} Federal Research Center ``Computer Science and Control'' of Russian Academy of Sciences, Moscow, Russia}
\IEEEauthorblockA{\IEEEauthorrefmark{3} Smart Engines Service LLC, Moscow, Russia}
\IEEEauthorblockA{\IEEEauthorrefmark{4} Institute for Information Transmission Problems, RAS, Moscow, Russia}
Email: trusov.av@smartengines.ru, limonova@smartengines.com, dimonstr@iitp.ru, vva777@gmail.com}


\maketitle

\begin{abstract}
Low-bit quantized neural networks (QNNs) are of great interest in practical applications because they significantly reduce the consumption of both memory and computational resources. Binary neural networks (BNNs) are memory and computationally efficient as they require only one bit per weight and activation and can be computed using Boolean logic and bit count operations. QNNs with ternary weights and activations (TNNs) and binary weights and ternary activations (TBNs) aim to improve recognition quality compared to BNNs while preserving low bit-width. However, their efficient implementation is usually considered on ASICs and FPGAs, limiting their applicability in real-life tasks. At the same time, one of the areas where efficient recognition is most in demand is recognition on mobile devices using their CPUs. However, there are no known fast implementations of TBNs and TNN, only the daBNN library for BNNs inference. In this paper, we propose novel fast algorithms of ternary, ternary-binary, and binary matrix multiplication for mobile devices with ARM architecture. In our algorithms, ternary weights are represented using 2-bit encoding and binary – using one bit. It allows us to replace matrix multiplication with Boolean logic operations that can be computed on 128-bits simultaneously, using ARM NEON SIMD extension. The matrix multiplication results are accumulated in 16-bit integer registers. We also use special reordering of values in left and right matrices. All that allows us to efficiently compute a matrix product while minimizing the number of loads and stores compared to the algorithm from daBNN. Our algorithms can be used to implement inference of convolutional and fully connected layers of TNNs, TBNs, and BNNs. We evaluate them experimentally on ARM Cortex-A73 CPU and compare their inference speed to efficient implementations of full-precision, 8-bit, and 4-bit quantized matrix multiplications. Our experiment shows our implementations of ternary and ternary-binary matrix multiplications to have almost the same inference time, and they are 3.6 times faster than full-precision, 2.5 times faster than 8-bit quantized, and 1.4 times faster than 4-bit quantized matrix multiplication but  2.9 slower than binary matrix multiplication.  
\end{abstract}

\IEEEpeerreviewmaketitle

\section{Introduction}

Convolutional neural networks (CNNs) are the primary tool for solving various computer vision problems: pattern recognition~\cite{wen2021application}, detection of different objects~\cite{teplyakov2021line, ye2020building}, semantic segmentation~\cite{kurnikov2021dnns} and many others. 
Although new transformer-based~\cite{dosovitskiy2020image, liu2021swin} or deep MLP-based~\cite{tolstikhin2021mlp, yu2022s2} neural networks sometimes outperform CNNs on challenging datasets, they are usually harder to train, have more parameters and require more computational resources for inference~\cite{yu2022s2, liu2021we}. 
That is why CNNs remain irreplaceable in practical applications.
The high performance of CNNs is essential for on-device intelligence systems, which allows for solving computer vision problems directly on a mobile device without the transmission of information to an external server and thus solve them faster, as well as in a more energy-efficient and secure way~\cite{gorospe2021generalization}.

The most computationally-challenging operation in a CNN is a discrete convolution of a feature map with a convolution kernel. 
A widely used approach for its efficient computation is a general matrix multiplication-based (GeMM-based) approach. Using this approach, the feature map and the convolution kernel are unrolled to matrices and then multiplied with the help of optimized BLAS libraries~\cite{chellapilla2006high}. The most common method for transforming a feature map to a matrix is the im2col method. Unfortunately, it suffers from significant memory overhead, which is why several more resource-efficient methods were proposed~\cite{anderson2020high, trusov2021pim2col}. 


It is worth mentioning that efficient algorithms for discrete convolution are not limited to GeMM-based.
For example, FPGAs, ASICs, and sometimes GPUs can benefit from the reduction of the number of multiplications using Winograd's minimal filtering algorithms~\cite{lavin2016fast} but also can use more straightforward algorithms~\cite{sotiropoulos2021porting}.
At the same time, on CPUs it is critical to optimize data flow in memory, so that the number of cache misses is low and data-parallel execution with Single Instruction Multiple Data (SIMD) instructions is possible.
That can be achieved with the help of just-in-time code generation of direct convolution for specific kernel sizes~\cite{georganas2018anatomy}. 
However, all mentioned algorithms are specific to devices and/or convolution parameters, so GeMM-based algorithms are still widely used in practice.

One of the most efficient ways to speedup and reduce the memory footprint of a CNN is to replace floating-point values of weights and activations with integers. This process is called quantization, and a neural network with integer weights is referred to as quantized neural network (QNN)~\cite{gorospe2021generalization}. Widely used 8-bit quantization allows for a 4-times reduction of network size and significant speedup on mobile CPUs while maintaining the quality close to full precision models~\cite{jacob2018quantization}.
4-bit QNNs demonstrate a noticeable drop in recognition quality on challenging tasks~\cite{banner2018post, choukroun2019low}; still, 4-bit quantization can be used to accelerate CPU inference of small CNNs significantly~\cite{trusov2021fast}.
The most memory-efficient quantization is binarization: in binary QNNs (BNNs), weights and activations only take the values of $1$ or $-1$ and require a single bit for storage. 
In BNNs, convolutions and matrix multiplications can be computed using only XOR/XNOR and bit count operations~\cite{rastegari2016xnor}.
That makes such networks exceptionally computationally efficient, especially on FPGAs and ASICs. 
There is also a CPU implementation of BNNs available in daBNN library~\cite{zhang2019dabnn}. 

Although training techniques for binary networks have improved in the past few years~\cite{bulat2019xnor, tang2017train}, they still show a significant gap in accuracy in comparison with full-precision ones.
Ternary neural networks (TNNs) allow weights and activations take the values $1$, $0$ or $-1$~\cite{alemdar2017ternary}. 
They show higher quality than BNNs and can be efficiently implemented on ASICS and FPGAs~\cite{prost2017scalable, deng2018gxnor}.
Ternary-binary networks (TBNs), which have ternary activations and binary weights, take medium place between TNNs and BNNs in terms of computational complexity and show almost the same recognition quality as TNNs \cite{wan2018tbn}. However, no computationally-efficient CPU-oriented algorithms of ternary and ternary-binary convolution and/or matrix multiplications were previously published to the best of our knowledge. 

In this paper, we propose novel algorithms for high-performance matrix multiplication of binary, ternary, and ternary-binary matrices for CPUs with ARMv8 architecture. 
They can be used in convolutional and fully-connected (linear) layers of BNNs, TNNs, and TBNs to obtain computationally efficient inference of such networks on mobile devices. 
Our algorithms use binary logic operations instead of multiplications and accumulate their products in 16-bit integer values. 
It allows us to take full advantage of data-parallel computing with the help of SIMD extension NEON of ARM CPUs. 

In the experimental section of our work, we compare the performance of the proposed algorithms to computationally-efficient algorithms of matrix multiplication for different data types: 32-bit floating-point, 8-bit integer from Google's gemmlowp library \cite{jacob2017gemmlowp}, 4-bit presented in \cite{trusov2021fast}, and binary from daBNN library \cite{zhang2019dabnn}.

\section{Efficient matrix multiplication on CPUs}
\subsection{High-performance GeMM}
  \label{sec:gemm}
  In this section, we discuss commonly used high-performance matrix multiplication computation methods. 
  We use $A$ to denote left matrix of size $m  \times k$, $B$ to denote right matrix of size $ k \times n$ and $C = AB$ to denote their product. 
  We refer to $m$, $n$ and $k$ as ``height'', ``width'' and ``depth'' of matrix multiplication respectively. 
  $X_{ij}$ denotes element of $X$ in row $i$ and column $j$.
  
  Most modern machine-learning and linear algebra libraries implement matrix multiplication in the following way: they split the matrix $A$ along rows, the matrix $B$ along columns, and possibly both along the depth, then they use a high-performance function called ``inner-kernel'' or ``microkernel'' to compute small block of matrix $C$. 
  Actually, there are several ways in which matrices $A$, $B$, and $C$ could be split as described in detail in \cite{goto2008anatomy}.
  One of them is presented as Algorithm~\ref{algo:gemm1}. 
  Here matrix $A$ is first split by blocks of $m_{blk}$ rows, values in those blocks are reordered by function \texttt{PackNRowsA} so that small blocks of its rows can be easily processed by microkernel, the result is stored in $A_{buf}$.
  Likewise matrix $B$ is split by blocks of $n_{blk}$ columns, reordered by \texttt{PackNColsB} and stored into $B_{buf}$.
  After that algorithm extracts smaller blocks: $m_{mk}$ rows and $k_{eff} \leq k_{blk}$ columns from $A_{buf}$, and $n_{mk}$ columns and  $k_{eff}$ rows from $B_{buf}$.
  Those blocks are multiplied by the microkernel.
  Values of $m_{mk}$ and $n_{mk}$ as well as the storage order of values in buffers $A_{buf}$ and $B_{buf}$ depend on the microkernel implementation.
  Values of $k_{blk}$, $m_{blk}$ and $n_{blk}$ are independent of the microkernel. They are chosen so that reordered buffers fit into the L2 CPU cache and smaller buffers that microkernel processes fit into the L1 cache.
  This way, the number of cache misses is minimized, so multiplication computes faster.
  
   \begin{algorithm}[h!]
    \small
        \caption{General high-level GeMM algorithm}
        \label{algo:gemm1}
        \KwIn{
        $m$, $n$, $k$ -- height, width, depth; \newline
        $A$ -- left matrix $m \times k$; \newline
        $B$ -- right matrix $k \times n$;
        }
        \KwOut{
        $C$ -- matrix $m \times n$, $C = AB$
        }
        \SetKwFunction{mk}{microkernel}
        \SetKwData{ablk}{$A_{buf}$}
        \SetKwData{bblk}{$B_{buf}$}
        \SetKwData{apan}{$A\_block$}
        \SetKwData{bpan}{$B\_block$}
        \SetKwFunction{pka}{PackNRowsA}
        \SetKwFunction{pkb}{PackNColsB}
       
        \For{$y \gets 0 $; $y < m$; $y \gets y + m_{blk}$}{
            $m_{eff} \gets \min(m_{blk}, m - y)$ 
            
            \ablk$[0 : m_{eff} - 1]$ $\gets$ \pka{$A$, y, $m_{eff}$}
            
            \For{$x \gets 0 $; $x < n$; $x \gets x + n_{blk}$}{
            
                $n_{eff} \gets \min(n_{blk}, n - x)$

                \bblk$[0 : n_{eff} - 1]$ $\gets$ \pkb{$B$, x, $n_{eff}$}
                
                \For{$d \gets 0 $; $d < k$; $d \gets d + k_{blk}$}{
                
                    $k_{eff} \gets \min(k_{blk}, k - d)$
                    
                    \For{$c \gets 0 $; $c < n_{eff}$; $c \gets c + n_{mk}$}{
                    
                        \For{$r \gets 0 $; $r < m_{eff}$; $r \gets r + m_{mk}$} {
                            
                            $j \gets y + c;~ i \gets x + r$
                        
                            $C[j: j + m_{mk} - 1][i : i + n_{mk} - 1]$ $\gets$ \mk{\\
                            \hspace*{2em}\ablk$[r : r + m_{mk} - 1]$,\\
                            \hspace*{2em}\bblk$[c : c + n_{mk} - 1]$,~~$k_{eff}$}
                            
                        }
                    }
                    
                }
                
            }
            
        }
  \end{algorithm}
  
  If we consider neural network inference, the matrix $B$ is a weights matrix. 
  It is usually small enough to fit into the L2 cache.
  Also, it does not change during inference of the neural network, so one can reorder it and store it in the \texttt{PackedB} buffer beforehand.
  Taking it into consideration, we use a little simpler algorithm to compute convolutions (Algorithm~\ref{algo:gemm2}).
  In our version of the algorithm, buffer $A_{buf}$ is noticeably smaller: it contains only $m_{mk}$ rows and $k_{eff} \leq k_{blk}$ columns from the matrix $A$. That can help the inference on mobile devices where memory is limited.
  
  \begin{algorithm}[h!]
    \small
        \caption{Our high-level GeMM algorithm}
        \label{algo:gemm2}
        \KwIn{
        $m$, $n$, $k$ -- height, width, depth; \newline
        $A$ -- left matrix $m \times k$; \newline
        \texttt{PackedB} -- pre-reordered right matrix $k \times n$;
        }
        \KwOut{
        $C$ -- matrix $m \times n$, $C = AB$
        }
        \SetKwFunction{mk}{microkernel}
        \SetKwData{ablk}{$A_{buf}$}
        \SetKwData{bblk}{\texttt{PackedB}}
        \SetKwFunction{pka}{PackNRowsA}

        \For{$d \gets 0 $; $d < k$; $d \gets d + k_{blk}$}{
        
            $k_{eff} \gets \min(k_{blk}, k - d)$
            
            \For{$r \gets 0 $; $r < m$; $r \gets r + m_{mk}$} {
                
                \ablk$[0 : m_{eff} - 1]$ $\gets$ \pka{A, r, $m_{mk}$}
                
                \For{$c \gets 0 $; $c < n$; $c \gets c + n_{mk}$} {
                
                    $C[r : r + m_{mk} - 1][c : c + n_{mk} - 1]$ $\gets$ \mk{\ablk,\\
                    \hspace*{5em}\bblk$[c : c + n_{mk} - 1]$, $k_{eff}$}
                    
                }
            }
            
        }
  \end{algorithm}
  
  In Algorithms \ref{algo:gemm1} and \ref{algo:gemm2} height and width ($n$ and $m$) are assumed to be multiples of microkernel height and width ($m_{mk}$ and $n_{mk}$) respectively. 
  In practice, several microkernels of different shapes are implemented to compute multiplications on matrices of arbitrary shapes.
  Most computations are performed with a bigger microkernel, and smaller microkernels compute the remains.
  
  The values $m_{mk}$ and $n_{mk}$ should be chosen as high as possible to minimize the number of loads and stores from memory but low enough so that the result of matrix multiplication can be stored in CPU registers.
  This way, in the inner loop, the microkernel loads values from \texttt{Ablock} and \texttt{Bblock}, multiplies them, and accumulates results in registers.
  After whole multiplication is computed, the results are offloaded to memory.
  So, the shape of the microkernel depends on the number of registers available on specific CPU architecture and bit-width of values used in computation (the smaller it is, the more values can be stored in a single register).
  For example, AArch64 (ARMv8) CPUs have 32 128-bit SIMD registers.

\subsection{Integer GeMM}

  Now let us point out key differences between floating-point matrix multiplication in CNNs and integer matrix multiplication in QNNs. 
  The first is that CNNs just multiply matrices. 
  On ARM CPUs multiplication microkernel can use FMLA instruction that for 3 SIMD registers $a$, $b$ and $c$ (each holding four 32-bit floating-point values) computes $\texttt{FMLA}(a, b, c) =  a + b c$ element-wise.
  
  In QNNs, integer computations are used to approximate floating-point computations of CNNs.
  That can be done through linear quantization -- the procedure in which all floating-point values of a neural network layer weights or activation are approximated with integers according to: 
  
  \begin{equation}
    \hat{x} =  \max\Bigg(\min\Big(\floor*{\frac{x}{s} - z}, Q\Big), 0\Bigg),
    \label{eq:quant}
  \end{equation}
  where $Q$ is a maximal quantized value, $s$ is a floating-point value called ``scale'' and $0 \leq z < Q$ is an integer value called ``zero-point'' ($\hat{0} = z$), $\floor*{y}$ denotes integer part of $y$.
  For $n$-bit quantization $Q = 2^n - 1$. Strategies to obtain scale and zero-point can vary~\cite{jacob2017gemmlowp, choukroun2019low, banner2018post}.
  

  Let us consider $\hat{A}$ to be a quantized approximation of matrix $A$ with scale $s_A$ and zero-point $z_A$. We use similar notation for $\hat{B}$, $B$, $s_B$, and $z_B$. Then matrix multiplication can be approximated as:
  
  \begin{equation}
  \begin{split}
    C_{ij} &= \sum_{t = 1}^{k} A_{it}B_{tj} \simeq \sum_{t = 1}^{k} s_A(\hat{A}_{it} - z_A)s_B({\hat{B}_{tj}} - z_B) = \\ 
    &= s_A s_B\sum_{t = 1}^{k} \bigg((\hat{A}_{it} - z_A)({\hat{B}_{tj}} - z_B)\bigg) = s_A s_B \tilde{C}_{ij},
    \label{eq:mul}
  \end{split}
  \end{equation}
  where $\tilde{C}_{ij}$ can be computed using integer-only arithmetic. In gemmlowp~\cite{jacob2017gemmlowp} and~\cite{trusov2021fast} it is done with the following transformation:
  \begin{equation}
  \begin{split}
    \tilde{C}_{ij} &= \sum_{t = 1}^{k} \bigg((\hat{A}_{it} - z_A)(\hat{B}_{tj} - z_B)\bigg) = \sum_{t = 1}^{k} (\hat{A}_{it} \hat{B}_{tj}) -\\ 
    &- z_B \sum_{t = 1}^{k} \hat{A}_{it} - z_A \sum_{t = 1}^{k} \hat{B}_{tj} + k z_A z_B.
    \label{eq:quant_mul}
  \end{split}
  \end{equation}
  
  The first term of (\ref{eq:quant_mul}) presents matrix multiplication of quantized matrices: 8-bit with 32-bit product in case of gemmlowp and 4-bit with 16-bit product in case of~\cite{trusov2021fast}. 
  The second and third terms do not depend on $j$ and $i$ respectively, so they are easier to compute: in terms of algorithmic complexity, the first term requires $O(m n k)$, the second -- $O(m k)$, the third -- $O(n k)$, and the fourth -- $O(1)$ operations.
  
  It is worth mentioning that integer accumulators can overflow, which limits the depth of matrix multiplication. 
  If matrices $A$ and $B$ hold $p$-bit values, and their product is accumulated in $q$-bit accumulators, then maximum depth that guarantees the absence of overflow is:
  
  \begin{equation}
    \label{eq:k_max}
    k_{\max} = \floor*{\frac{2^q - 1}{(2^p - 1)^2}}.
  \end{equation}
  
  In GeMM-based convolution it limits the number of channels in the input feature map \cite{trusov2021fast}. Let us consider convolution with $H_k \times W_k$ kernel. Then the maximum number of channels in the input feature map, for which the absence of overflow in guaranteed is:
  
  \begin{equation}
    \label{eq:cin_max}
    C_{in\_\max} = \floor*{\frac{k_{\max}}{H_k W_k}}.
  \end{equation}

\section{Low-bit matrix multiplication}
  
  In this section we present our algorithms for matrix multiplication  of binary, ternary and ternary-binary matrices.
  
  We consider matrix multiplication $A \times B = C$ for three cases:
  \begin{description}
    \item[\textbf{BNN}] $A$ and $B$ are binary ($A_{ij} \in \{-1, 1\}$, $B_{ij} \in \{-1, 1\}$),
    \item[\textbf{TNN}] $A$ and $B$ are ternary ($A_{ij} \in \{-1, 0, 1\}$, $B_{ij} \in \{-1, 0, 1\}$),
    \item[\textbf{TBN}] $A$ is ternary and $B$ is binary ($A_{ij} \in \{-1, 0, 1\}$, $B_{ij} \in \{-1, 1\}$).
  \end{description}
  For all the cases we assume that $C$ holds integer values stored in signed 16-bit representation.

\subsection{Values, encoding and multiplication}
\label{sec:encode}
  
  In the considered algorithms values of matrix elements of $A$ and $B$ are either binary or ternary. For binary values we use single-bit encoding $x \to x^b: 1 \to 0, -1 \to 1$. This way matrix multiplication $z = x y$ can be computed using our representation as $z^b = x^b \oplus y^b$ (where $\oplus$ is an addition modulo 2 or XOR operation). Using this representation we can compute matrix multiplication using only XOR, bit-count and addition operation in the inner loop:
  
  \begin{equation}
      c =  \sum_{t = 1}^{k} a_t b_t = \sum_{t = 1}^{k} (1 - 2a^b_{t} \oplus b^b_{t}) = k - 2\sum_{t = 1}^{k}(a^b_{t} \oplus b^b_{t}).
      \label{eq:bin_mul}
  \end{equation}
 
 
 For ternary values we use 2-bit encoding $x \to (x^+, x^-)$: $1 \to (1, 0)$, $0 \to (0, 0)$, $-1 \to (0, 1)$, and code $(1, 1)$ is invalid.
 Using those operations we can compute ternary multiplication $z = x y$ as 
 $$(z^+, z^-) = ((x^+ \land y^+)\lor (x^- \land y^-) , (x^+ \land y^-)\lor(x^- \land y^+))$$
 and ternary-binary multiplication as 
 $$(z^+, z^-) = ((x^+ \lor y^b)\land(x^- \lor \overline{y^b}), (x^+ \lor \overline{y^b})\land(x^- \lor y^b)),$$ 
 where $\land$, $\lor$, and $\overline{*}$ denote logical AND, OR, and NOT.
 Truth tables for ternary and ternary-binary multiplication illustrating these equations are presented in Table \ref{tab:tnn_mul}.

 \begin{table}[ht]
    \caption{Ternary multiplication $z = x y$ and 
            Ternary-binary multiplication $u = x y$}
    \label{tab:tnn_mul}
    \begin{center}
    \begin{tabular}{|P{0.28cm}|P{0.28cm}|P{0.28cm}|P{0.28cm} P{0.28cm}|P{0.28cm}P{0.28cm}|P{0.28cm}|P{0.28cm}P{0.28cm}|P{0.28cm}P{0.28cm}|}
        \hline
        \hline
        \textbf{$x$} & \textbf{$y$} & \textbf{$z$} & \textbf{$x^+$} & \textbf{$x^-$} & \textbf{$y^+$} & \textbf{$y^-$}& \textbf{$y^b$} & \textbf{$z^+$} & \textbf{$z^-$} & \textbf{$u^+$} & \textbf{$u^-$}\\
        \hline
         1 &  1 &  1 & 1 & 0 & 1 & 0 & 0 & 1 & 0 & 1 & 0\\
         1 &  0 &  0 & 1 & 0 & 0 & 0 &-- & 0 & 0 & \multicolumn{2}{c|}{---}\\
         1 & -1 & -1 & 1 & 0 & 0 & 1 & 1 & 0 & 1 & 0 & 1 \\
         0 &  1 &  0 & 0 & 0 & 1 & 0 & 0 & 0 & 0 & 0 & 0 \\
         0 &  0 &  0 & 0 & 0 & 0 & 0 &-- & 0 & 0 & \multicolumn{2}{c|}{---}\\
         0 & -1 &  0 & 0 & 0 & 0 & 1 & 1 & 0 & 0 & 0 & 0 \\
        -1 &  1 & -1 & 0 & 1 & 1 & 0 & 0 & 0 & 1 & 0 & 1 \\
        -1 &  0 &  0 & 0 & 1 & 0 & 0 &-- & 0 & 0 & \multicolumn{2}{c|}{---}\\
        -1 & -1 &  1 & 0 & 1 & 0 & 1 & 1 & 1 & 0 & 1 & 0 \\
        \hline
        \hline
    \end{tabular}
    \end{center}
 \end{table}
 
 
 Assuming that $a_t b_t = (c^+, c^-)$ is a ternary or ternary-binary multiplication of a couple of elements as shown above, we can compute dot product (and matrix multiplication) as:
 
 \begin{equation}
      c =  \sum_{t = 1}^{k} a_t b_t = \sum_{t = 1}^{k} ((a_t, b_t)^+ - (a_t, b_t)^-). 
      \label{eq:tnn_mul}
  \end{equation}

\subsection{Binary microkernel}
  
  Now that we have encoding and operation definitions for matrix multiplication based on binary-logic operations available in the ARM instruction set, we move to matrix multiplication microkernels.
  As we stated in Section \ref{sec:gemm} microkernel multiplies $m_{mk}$ rows stored in buffer \texttt{Ablock} by $n_{mk}$ cols stored in buffer \texttt{Bblock}.
  To describe a multiplication microkernel we need to specify its shape $m_{mk} \times n_{mk}$, storage order in buffers \texttt{Ablock} and \texttt{Bblock} and operations used in the computation.
  All the microkernels we propose in this work have $16 \times 8$ shape and use 16 128-bit SIMD registers ($c00, ... c07, c10, ... c17$) to hold the corresponding block of matrix $C$ as 16-bit integers.
  However, storage order and SIMD instructions are different for all multiplications (BNN, TNN and TBN). 
  Let us start with binary matrix multiplication.
  
  To pack values into \texttt{Ablock} we
  \begin{enumerate}
      \item take 16 rows of $A$ matrix encoded as binary values and pack them into 8-bit values (each holding 8 consecutive bits from the corresponding row);
      \item store this 8-bit matrix in column-major order (first go 8 bits from the $1^{st}$ row, then 8 bits from the $2^{nd}$ and so on until the $16^{th}$, after that bits $9...16$ from the $1^st$ row, then bits $9...16$ from the $2^{nd}$ and so on).
  \end{enumerate}
  
  To pack values into \texttt{Bblock} we
  \begin{enumerate}
      \item take 8 columns of $B$ matrix encoded as binary values and pack them into 8-bit values (each holding 8 consecutive bits from the corresponding column);
      \item store this 8-bit matrix in row-major order (first 8 bits from the $1^{st}$ column, then 8 bits from the $2^{nd}$ and so on until the $8^{th}$, after that bits $9...16$ from the $1^{st}$ column, than bits $9...16$ from the $2^{nd}$ and so on).
  \end{enumerate}
  
\begin{figure}[ht]
  \centering
    \includegraphics[width=0.7\linewidth]{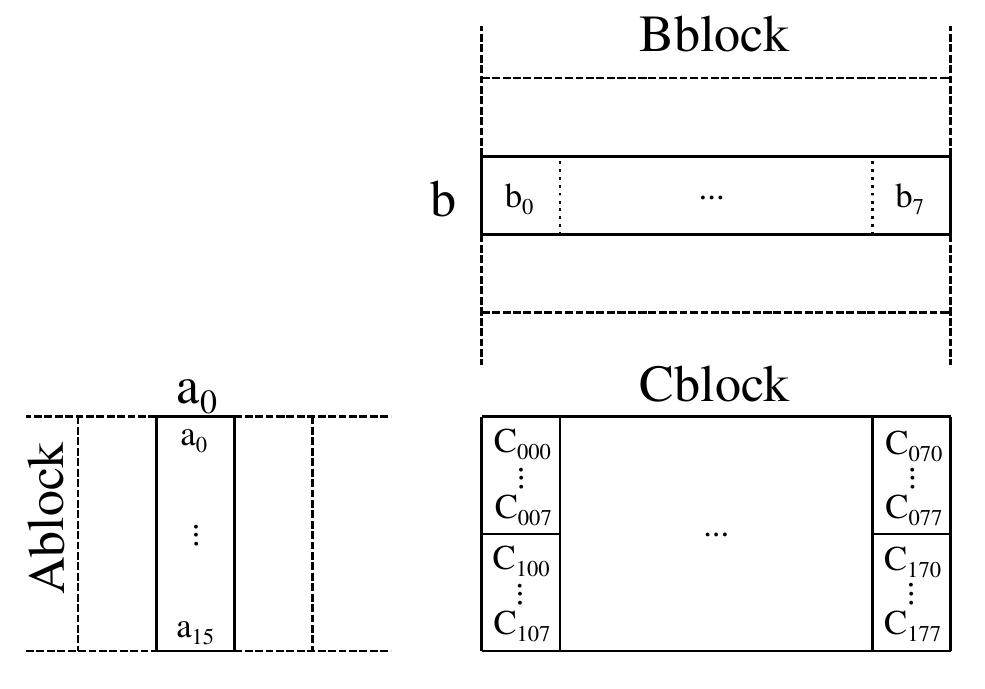}
    \caption{Binary GeMM microkernel.}
    \label{fig:bnn_ker}
  \end{figure}

  In one iteration of the loop over depth dimension in the microkernel (shown in Fig.~\ref{fig:bnn_ker}) we
  \begin{enumerate}
      \item load a column of 8-bit values from \texttt{Ablock} into two 128-bit registers $a$;
      \item load a row 8-bit values from \texttt{Bblock} into one 64-bit register $b$;
      \item for each 8-bit element of $b$ compute XOR (EOR instruction in ARM) with register $a$, count number of 1-bites in the ``product'' with CNT instruction and accumulate the result with SADDW instruction.
  \end{enumerate}

\subsection{Ternary microkernel}
  As described in Section~\ref{sec:encode} we use 2-bit representation for ternary values. To pack them into \texttt{Ablock} we
  \begin{enumerate}
      \item consider that $A^+$ and $A^-$ are stored as two separate matrices;
      \item take 8 rows of $A^+$ matrix and pack them into 8-bit values (each holding 8 consecutive bits from the corresponding row);
      \item do the same for $A^-$;
      \item store the first 8 8-bit elements from the first column of $A^+$ block, then store the first 8 8-bit elements from the first column of  $A-$ block, then the last 8 elements from the first columns from $A^+$ and $A^-$ blocks, then repeat this for the $2^{nd}$, the $3^d$ and all the remaining columns.
  \end{enumerate}
  
  To pack values into \texttt{Bblock} we
  \begin{enumerate}
      \item consider tat $B^+$ and $B^-$ are stored as two separate matrices;
      \item take 8 columns of $B^+$ matrix and pack them into 8-bit values (each holding 8 consecutive bits from the corresponding column);
      \item do the same for $B^-$;
      \item for the first row: store the $1^st$ element from $B^+$ block, then store the $1^st$ element from $B^-$ block, then the $2^{nd}$ elements and so on; repeat that procedure for all rows.
  \end{enumerate}
  
\begin{figure}[ht]
  \centering
    \includegraphics[width=0.7\linewidth]{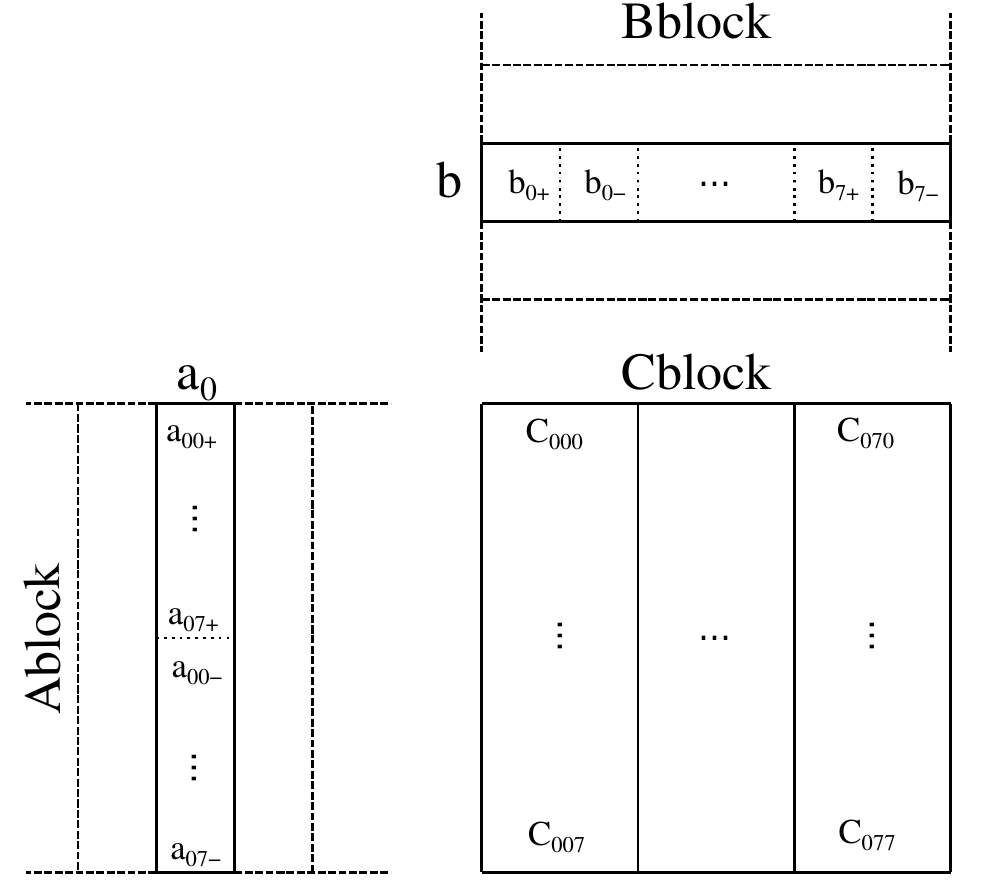}
    \caption{Ternary GeMM microkernel.}
    \label{fig:tnn_ker}
  \end{figure}

  On each iteration ternary microkernel loads a column from the \texttt{Ablock} as two 128-bit registers ($a0$ and $a1$) and a row from \texttt{Bblock} as a single 128-bit register $b$. For each pair of elements in $b$ it computes the product with registers $a0$ and $a1$ using AND and OR operations, then applies bit-count CNT to compute sums of ``+'' and ``-'' bits of the product, computes their difference with SSUBL and accumulates the result in the corresponding register with ADD. Fig~\ref{fig:tnn_ker} demonstrates a half of this microkernel -- 8 rows from \texttt{Ablock}.

\subsection{Ternary-binary microkernel}
  For ternary-binary multiplication we store values ternary in \texttt{Ablock} in the same way as for ternary multiplication and binary values in \texttt{Bblock} in the same way as for binary multiplication.
  
  This way multiplication this microkernel is almost the same as ternary with only two differences:
  \begin{itemize}
      \item \texttt{Bblock} contains 8 column of 8-bit values, so we use 64-bit register $b$ to load them;
      \item as we show in Section~\ref{sec:encode}, different instructions (OR, AND and ORN -- OR with negation of the second operand) are used to compute the product.
  \end{itemize}

  \begin{figure}[ht]
  \centering
    \includegraphics[width=0.7\linewidth]{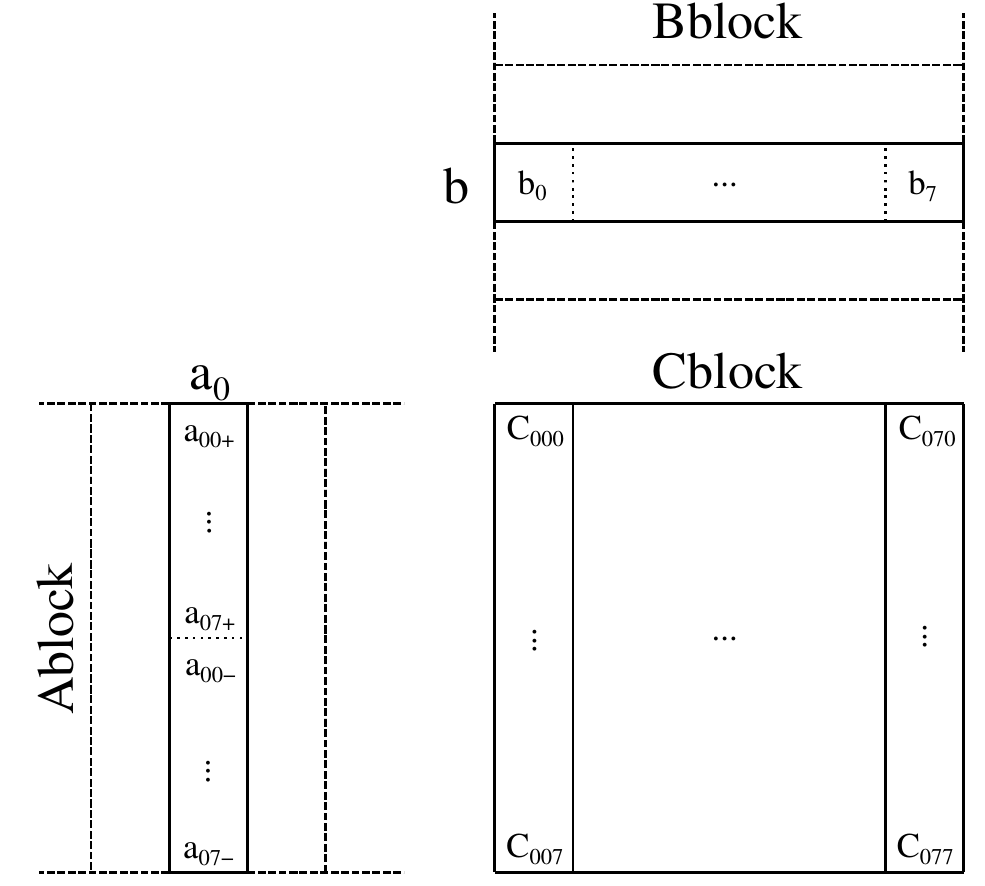}
    \caption{Ternary-binary GeMM microkernel.}
    \label{fig:tbn_ker}
  \end{figure}

\section{Matrix multiplication evaluation}
  In this section, we demonstrate the efficiency of the proposed ternary (TNN), ternary-binary (TBN), and binary (BNN) matrix multiplication on ARM Aarch64 CPUs and compare them to known efficient algorithms: binary from daBNN library~\cite{zhang2019dabnn} (daBNN), 8-bit from gemmlowp library~\cite{jacob2017gemmlowp} (U8), 4-bit from~\cite{trusov2021fast} with a microkernel upscaled to $24 \times 8$ size (U4, the original size was $24 \times 4$ for ARMv7 architecture), and our implementation of floating-point 32-bit baseline which uses the same register layout as gemmlowp, but computes operations in floating-point (F32). 

\subsection{Theoretical evaluation}
 
  In Table \ref{tab:com_th} we show a comparison of the microkernels for multiplication algorithms under consideration.
  We compare them by a number of columns in \texttt{Ablock} ($m$), a number of rows in \texttt{Bblock} ($n$), a step over depth per iteration~($k$), a number of computational instructions per iteration (\textbf{COM} -- FMLA for F32 algorithm, UMLAL/UMLAL2 for U8, etc.), a number of loads of SIMD register per microkernel instruction (\textbf{LD}), a number of other SIMD instructions per iteration (\textbf{MOV} -- MOV, DUP, INS, etc.), and by a number of SIMD instructions per microkernel element $\text{\textbf{INS}} = (\text{COM} + \text{LD} + \text{MOV}) / (n m k)$.
  We also estimate a maximum depth of multiplication ($k_{\max}$) as (\ref{eq:k_max}) for U8 and U4 algorithms. 
  In a ternary and binary matrix multiplication $x y = z, |z| \leq 1$, so $k_{\max}$ equals to the maximum possible value that a register can hold.
  TNN, TBN, and BNN use 16-bit signed registers, and $k_{\max} = 2^{15} - 1$. daBNN uses 32-bit floating point registers (with 23-bit significand field) to store \texttt{Cblock}, so $k_{\max} = 2^{23} - 1$.

  \begin{table}[ht]
    \caption{Comparison of the multiplication microkernels}
    \label{tab:com_th}
    \begin{center}
    \begin{tabular}{|c|c|r|r|r|r|r|}
        \hline
        \textbf{Algo} & $m \times n \times k$ & \textbf{COM} & \textbf{LD} & \textbf{MOV} & \textbf{INS} &  \textbf{$k_{\max}$}  \\
        \hline
        F32            & $12 \times 8 \times 1$  & 24  & 5  & 0  & 0.302 & ---\\
        U8             & $12 \times 8 \times 2$  & 48  & 5  & 5  & 0.302 & 66051\\
        U4             & $24 \times 8 \times 2$  & 48  & 5  & 16 & 0.180 & 291  \\
        \textbf{TNN}   & $16 \times 8 \times 8$  & 96  & 3  & 64 & 0.159 & 32767\\
        \textbf{TBN}   & $16 \times 8 \times 8$  & 96  & 3  & 56 & 0.151 & 32767\\
        \textbf{BNN}   & $16 \times 8 \times 8$  & 32  & 2  & 8  & 0.041 & 32767\\
        daBNN          & $8 \times 6 \times 128$ & 156 & 12 & 36 & 0.033 & 8388607\\
        \hline
    \end{tabular}
    \end{center}
 \end{table}

 A multiplication algorithm reaches maximal efficiency when multiplication parameters height, width, and depth are multiples of corresponding microkernel parameters ($m, n, k$).
 It limits the applicability of the multiplication algorithm in CNNs. 
 If a convolution is computed using im2col transformation, the height is the number of pixels in the input feature map; the width is the number of filters (from only a few in upper layers of small CNNs to hundreds and thousands in lower layers of big CNNs).
 $k_{\max}$ limits the number of input channels in the feature maps (\ref{eq:cin_max}).
 Taking all the limitations into account, we can see that U4 algorithm is only suitable for small CNNs, daDNN, on the other hand, will show better results in large networks, and all the rest (including three proposed algorithms) will work fine in small, medium and large CNNs.
 
 \subsection{Experimental evaluation}
 
 Although the number of instructions (\textbf{COM}, \textbf{LD}, \textbf{MOV}, and \textbf{INS} in Table \ref{tab:com_th}) can give us general ideas on which algorithm should have better computational efficiency, in practice the efficiency also depends on cache misses (which in turn depends on the order of loads and stores in memory), the ability of CPU to use the instruction pipelining (that depends on the order in which instructions are fetched).
 Furthermore, the overall efficiency of the matrix multiplication is affected also by reordering operations that prepare matrix blocks for microkernel and by post-processing in algorithms U8 and U4, which is shown in (\ref{eq:quant_mul}).
 That is why we experimentally measure the efficiency of all the algorithms under consideration.
 
 We implemented matrix multiplication algorithms F32, U4, TNN, TBN, TNN according to Algorithm \ref{algo:gemm2}. 
 All the microkernels were written on ARMv8 assembly to optimize the usage of SIMD registers.
 We ran time measurements for different values of height ($H \in \{72, 120, 240, 360\}$), width ($W \in \{24, 48, 72, 96\}$) and depth ($D \in \{128, 256, 384, 512\}$).
 Those values are chosen to be multiples of the microkernel size for each algorithm, so that they all can show the maximum efficiency. 
 They also are representative for matrix multiplications in small and medium CNNs, which can be used in real-life tasks on mobile CPUs.
 
We  ran  our  experiment  on  the  ARM  Cortex-A73  CPU, a part of Odroid-N2 development board with Linux.
 For each value of parameters, we took the median of 5 measurements (to exclude random errors) and repeated the whole experiment 50 times, taking the average of the measurements to reach 0.8\% empirical relative error of running time.
 We summarize them in Table \ref{tab:com_ex}. 
 Each cell compares algorithms $A$ and $B$ as $\mathbf{E}_{\theta}(T_B(\theta) / T_A(\theta)) $, where $T_A$ and $T_B$ denote execution times of corresponding on test $\theta$, $\mathbf{E}$ is mathematical expectation.
 
 \begin{table}[ht]
    \caption{Efficiency ratio of the multiplication algorithms}
    \label{tab:com_ex}
    \begin{center}
    \begin{tabular}{|c|c|c|c|c|c|c|c|}
        \hline
        \diagbox{B}{A}  & F32 & U8 & U4 & \textbf{TNN} & \textbf{TBN} & \textbf{BNN} & daBNN\\
        \hline
        F32             & 1.00 & 1.44 & 2.52 & 3.63 & 3.75 & 10.9 & 9.60\\
        \hline
        U8              & 0.69 & 1.00 & 1.75 & 2.51 & 2.60 & 7.52 & 6.63\\
        \hline
        U4              & 0.40 & 0.57 & 1.00 & 1.44 & 1.49 & 4.32 & 3.81\\
        \hline
        \textbf{TNN}    & 0.28 & 0.40 & 0.70 & 1.00 & 1.03 & 2.99 & 2.64\\
        \hline
        \textbf{TBN}    & 0.27 & 0.39 & 0.67 & 0.97 & 1.00 & 2.90 & 2.55\\
        \hline
        \textbf{BNN}    & 0.093 & 0.13 & 0.23 & 0.34 & 0.35 & 1.00 & 0.88\\
        \hline
        daBNN           & 0.11 & 0.15 & 0.27 & 0.39 & 0.40 & 1.15 & 1.00\\
        \hline
    \end{tabular}
    \end{center}
 \end{table}
 
 According to Table \ref{tab:com_ex} our TNN algorithm significantly outperforms matrix multiplication for types with greater bit-width: it is 3.6 faster than F32, 2.5 times faster than U8 and 1.4 times faster than U4.
 TBN is only slightly faster than TNN because of a simpler data flow in \texttt{Bblock}.
 BNN is almost 3 times faster than TNN (and 2.9 times faster than TBN. 
 Moreover our implementation of binary multiplication turns out to be 1.15 times faster than that of daBNN library, due to a bigger microkernel and 16-bit representation of the result.  
 
    Proposed algorithms for ternary, ternary-binary, and binary matrix multiplication show significantly higher computational efficiency than algorithms with greater bit-width.
    They can be used for the inference of low-bit CNNs on mobile devices and do not pose strict constraints on the network architecture (although to achieve maximal efficiency, numbers of channels in feature maps and convolutional filters should be multiples of 8).
    It opens an opportunity to investigate the trade-off between recognition quality (which usually decreases along with the bit-width of QNN) and efficiency gain from low-bit quantization of several layers or whole CNNs.
    Note that in the pursuit of high computational efficiency, different libraries for network inference usually implement direct convolution algorithms for the most common shape of convolution kernels that do not rely on matrix multiplication. 
    For example, daBNN library implements $3\times3$ binary convolution directly.
    Our ideas of encoding and computation of ternary and binary dot products can be used in those algorithms as well.
    
\section{Conclusion}
    In this paper, we presented algorithms for matrix multiplication of ternary, ternary-binary, and binary matrices of ARM CPUs. 
    Those algorithms are superior to existing CPU implementations of matrix multiplication. 
    For example, ternary, ternary-binary, and binary multiplications are 3.6, 3.7, and 11 times faster, respectively, than floating-point multiplication.
    They have also shown good computational efficiency compared to 8-bit multiplication from Google's gemmlwop library (ternary multiplication is 2.5 times faster) and binary multiplication from daBNN library (our binary algorithm is 1.15 faster).
    
    Our algorithms can be used in the GeMM-based convolution implementations of CNNs over a wide range of parameters, which allows for computationally- and resource-efficient inference of low-bit CNNs on mobile devices.

\section*{Acknowledgment}
This work is partially supported by Russian Foundation for Basic Research (projects 19-29-09066 and 19-29-09092). 

\bibliographystyle{IEEEtran}
\bibliography{bibtex}

\end{document}